\documentclass[letterpaper, 10 pt, conference]{ieeeconf}  

\IEEEoverridecommandlockouts                              

\usepackage[symbol]{footmisc}
\usepackage{marvosym}
\usepackage{amsmath}
\usepackage{graphicx}
\usepackage{booktabs}
\usepackage{amssymb}
\usepackage{url}
\usepackage{hyperref}
\title{\LARGE \bf
Grounding Actions in Camera Space: Observation-Centric Vision-Language-Action Policy
}

\author{Tianyi Zhang$^{1, 2}$\authorrefmark{1}, Haonan Duan$^{3}$\authorrefmark{1}, Haoran Hao$^{4, 2}$, Yu Qiao$^{2}$, Jifeng Dai$^{5}$ and Zhi Hou$^{2}$\authorrefmark{2}  
\thanks{$^{1}$College of Computer Science and Technology, Zhejiang University}%
\thanks{$^{2}$Shanghai AI Lab}
\thanks{$^{3}$SenseTime Research}
\thanks{$^{4}$Nanjing University}
\thanks{$^{5}$Tsinghua University}
\thanks{\authorrefmark{1} Equal Contribution}
\thanks{\authorrefmark{2} Corresponding Author}
}

\begin{document}

\maketitle
\thispagestyle{empty}
\pagestyle{empty}
\begin{figure*}[!h]
      \centering
      \includegraphics[width=0.98\textwidth]{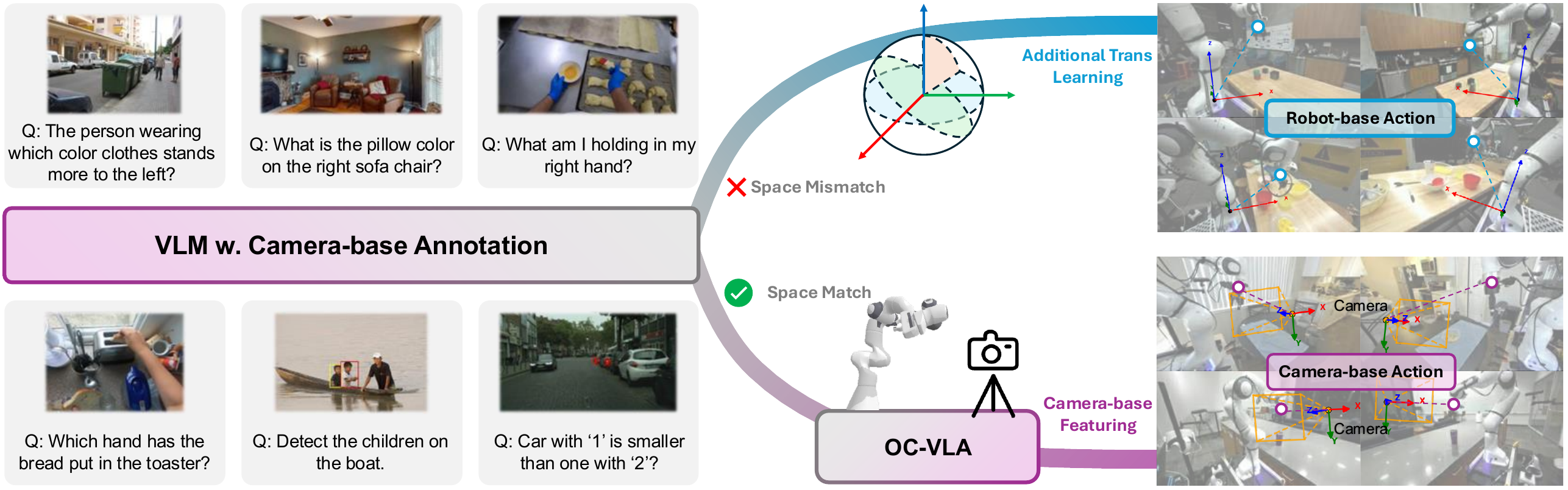}
      \caption{We introduce the Observation-Centric VLA (OC-VLA) framework. By transforming end-effector actions from the robot base coordinate to the third-person camera coordinate, OC-VLA aligns action predictions with visual observations across diverse viewpoints, enabling improved generalization and robustness in manipulation tasks.}
      \label{figure01}
   \end{figure*}

\begin{abstract}

Vision-Language-Action (VLA) models frequently encounter challenges in generalizing to real-world environments due to inherent discrepancies between observation and action spaces. Although training data are collected from diverse camera perspectives, the models typically predict end-effector poses within the robot base coordinate frame, resulting in spatial inconsistencies. To mitigate this limitation, we introduce the {\bf Observation-Centric VLA (OC-VLA)} framework, which grounds action predictions directly in the camera observation space. Leveraging the camera’s extrinsic calibration matrix, OC-VLA transforms end-effector poses from the robot base coordinate system into the camera coordinate system, thereby unifying prediction targets across heterogeneous viewpoints. This lightweight, plug-and-play strategy ensures robust alignment between perception and action, substantially improving model resilience to camera viewpoint variations. The proposed approach is readily compatible with existing VLA architectures, requiring no substantial modifications. Comprehensive evaluations on both simulated and real-world robotic manipulation tasks demonstrate that OC-VLA accelerates convergence, enhances task success rates, and improves cross-view generalization. The code will be publicly available.
\end{abstract}

\section{INTRODUCTION}

Inspired by the remarkable progress of multimodal large models, recent advances in vision-language-action (VLA) models~\cite{brohan2022rt,kim2024openvla,team2024octo,belkhale2024rt,black2024pi_0,hou2025dita} have focused on leveraging large-scale robot data from heterogeneous sources for pre-training, with the objective of enhancing generalization capabilities. Although this paradigm has achieved impressive performance across a variety of benchmarks, it remains fundamentally constrained by the intrinsic limitations of the robotics domain---namely, the relatively modest scale and high cost of data collection when compared to the web-scale corpora used in vision-language model (VLM) pre-training \cite{o2024open, walke2023bridgedata}. Consequently, the ability of current VLA models to generalize effectively in real-world environments remains limited, leaving substantial room for further advancement.

A common practice in VLA modeling is to adapt pretrained vision-language or vision encoders for downstream robotic tasks~\cite{brohan2022rt,o2024open,kim2024openvla,hou2025dita}. However, these vision models are primarily trained and supervised within the image or camera coordinate system, resulting in latent representations that are inherently aligned with camera viewpoints. In contrast, most robotic control signals are defined in the robot base coordinate system~\cite{brohan2022rt,o2024open,kim2024openvla,hou2025dita}. This discrepancy introduces a misalignment between the perception and action spaces, which can hinder effective policy learning, especially during the transfer of pretrained vision models to robotic control tasks.

Moreover, robot datasets are typically collected under diverse camera viewpoints and heterogeneous hardware configurations~\cite{o2024open, khazatsky2024droid,walke2023bridgedata}, where the robot base is not always within the camera's field of view. In such settings, the same action expressed in the robot base coordinate system must be inferred from different third-person camera views. This implicitly requires the model to reconstruct or reason about consistent 3D actions from limited 2D observations---a fundamentally ill-posed challenge when only single- or dual-view inputs are available. Predicting actions defined in the robot base coordinate system becomes even more challenging, as it necessitates an implicit understanding of the transformation between robot and camera spaces. Such inconsistencies are particularly detrimental during large-scale pre-training~\cite{brohan2023rt,kim2024openvla}, where diverse camera viewpoints are common: images capturing the same robot action from different angles are forced to share a single supervision signal in robot space, thereby introducing learning conflicts and hindering generalization.  


To address these issues, we propose a novel paradigm that decouples the end-effector action from the robot base coordinate system and instead predicts actions directly in the third-person camera coordinate system, named Observation-Centric VLA (OC-VLA). Specifically, given the extrinsic transformation between the robot base and each camera, we transform the robot-space end-effector actions into their equivalent representations in the camera coordinate frame and adopt these as prediction targets. By anchoring the action target in the same space as the observation (i.e., the image plane), this formulation alleviates the misalignment between perception and action modalities and mitigates the ambiguity introduced by camera viewpoint variations. Furthermore, it explicitly encourages the model to learn the relative spatial relationships between the robot and the cameras, thereby enhancing its capacity to generalize effectively across diverse viewpoints and hardware configurations.



The proposed approach is evaluated across both simulated environments and real-world robotic platform. Experimental results consistently demonstrate that employing camera-space end-effector actions as prediction targets yields substantial performance gains over baselines that operate in robot coordinates. Notably, our method exhibits markedly improved adaptability to previously unseen camera viewpoints, underscoring its strong potential for robust generalization in diverse real-world deployment scenarios.

\section{Related Work}

\subsection{Robotic Manipulation}

Robotic manipulation has wide applications in the real world, but still faces significant challenges in complex environments and tasks. 
Compared with traditional methods, learning-based manipulation has gained significant attention in recent years \cite{kroemer2021review}. A common strategy for learning to predict actions is reinforcement learning \cite{dalal2024plan-seq-learn, yamada2021motion, xia2020relmogen}.
Another approach is to provide offline expert demonstrations for supervised learning \cite{brohan2022rt, shridhar2022cliport, shridhar2023perceiver}, which trains the model to imitate the action performed by experts. However, both approaches are data-driven and sensitive to environmental changes, limiting their effectiveness in open-world applications.
Recently, the development of large language models (LLMs) and vision-language models (VLMs) has made reasoning and planning possible for solving complex tasks that require human knowledge \cite{li2024manipllm, jin2024robotgpt, singh2023progprompt}. However, due to limitations in their pretraining data, these models are still unable to control robots and address real-world tasks effectively.
Vision-Language-Action (VLA) models \cite{brohan2022rt,li2023vision, zhen20243d, kim2024openvla, hou2025dita, qu2025spatialvla, cheang2024gr, black2024pi_0} are trained on large-scale observation-action pairs and have strong capabilities in unified perception, reasoning, planning, and control, making them a promising solution for achieving unified robotic manipulation. Nevertheless, the generalization of current VLA models is limited, and the observation space action prediction is poorly investigated.

\subsection{Vision-Language-Action Model}

Vision-Language-Action (VLA) models~\cite{brohan2022rt,kim2024openvla,team2024octo,qu2025spatialvla,o2024open,belkhale2024rt,fang2023rh20t,driess2023palm,hou2025dita} have become the popular framework for generalist robot policies, enabling robots to interpret natural language instructions and visual observations for robust action generation. Recent advances leverage large-scale multi-modal backbones and foundation models~\cite{wu2023unleashing,cheang2024gr,li2025gr,huang2025enerverse,lynch2020language,reuss2023goal,ha2023scaling,myers2023goal,zhang2022language,chen2023playfusion,tian2024predictive}, improving generalization across tasks and embodiments. Diffusion models~\cite{ho2020denoising,rombach2022high,dhariwal2021diffusion,peebles2023scalable,videoworldsimulators2024} have shown strong performance in multi-modal action modeling~\cite{liang2024skilldiffuser,wang2024one,cao2024mamba,wang2024sparse,chen2024diffusion,chi2023diffusion,ze20243d,ke20243d,reuss2024multimodal,liu2024rdt,wen2025dexvla}, yet most existing approaches rely on U-Net or shallow cross-attention architectures, which limit scalability to more diverse tasks. To address complex scenarios, recent works integrate VLM embeddings with MLP diffusers~\cite{team2024octo,wen2024diffusion}, or utilize Transformer-based \cite{NIPS2017_transformer} decoders for bimanual and multi-modal manipulation~\cite{liu2024rdt,dasari2024ingredients}, further pushing the frontier of unified VLA policy learning. Although VLA models have made great progress, most of them rely on a specific observation space to predict actions. However, differences in environment setups during data collection make it difficult to use large-scale web data directly for training, which limits their performance. Meanwhile, current manipulation datasets encompass a wide range of camera views, whereas existing VLA approaches typically focus on action prediction based on the robot's base coordinates. The discrepancy between the coordinates of action prediction and observation poses a significant challenge for policy learning.

\section{Method}

In this section, we provide a detailed overview of OC-VLA, i.e., grounding actions in the observation (camera) space. We begin with the model structure and action modeling as preliminaries, followed by an introduction to the camera-centric action prediction approach. We then analyze the differences between camera-coordinate and robot-coordinate optimization. 

\subsection{Preliminary: Model Structure, Action Modeling} 
Vision-language-action (VLA) models have converged toward a common architectural pattern~\cite{team2024octo,kim2024openvla,black2024pi_0,hou2025dita}, where action prediction is built upon a vision-language backbone. Following this paradigm, we adopt a lightweight 300M VLA model~\cite{hou2025dita} for evaluation, which has demonstrated competitive performance using only a third-person camera image and language instructions as input. Specifically, we follow Dita~\cite{hou2025dita}, where the language instruction is encoded using a CLIP text encoder~\cite{radford2021learning}, and the third-person image is processed using DINOv2~\cite{oquab2023dinov2}. The image features are further selected and modulated by the language instruction via a Q-Former~\cite{li2023blip} equipped with FiLM~\cite{perez2018film} conditioning layers.

Current VLA models typically employ one of two types of action spaces for end-effector control: discrete action spaces~\cite{kim2024openvla,brohan2022rt} and continuous action spaces~\cite{team2024octo,black2024pi_0}. To thoroughly evaluate the effectiveness of our proposed approach, we conduct experiments on models using both types of action spaces. Based on the baseline architecture, we implement a variant specifically designed for discrete action prediction or continuous action prediction.
\begin{figure}[t]
\centering
\includegraphics[width=0.98\columnwidth]{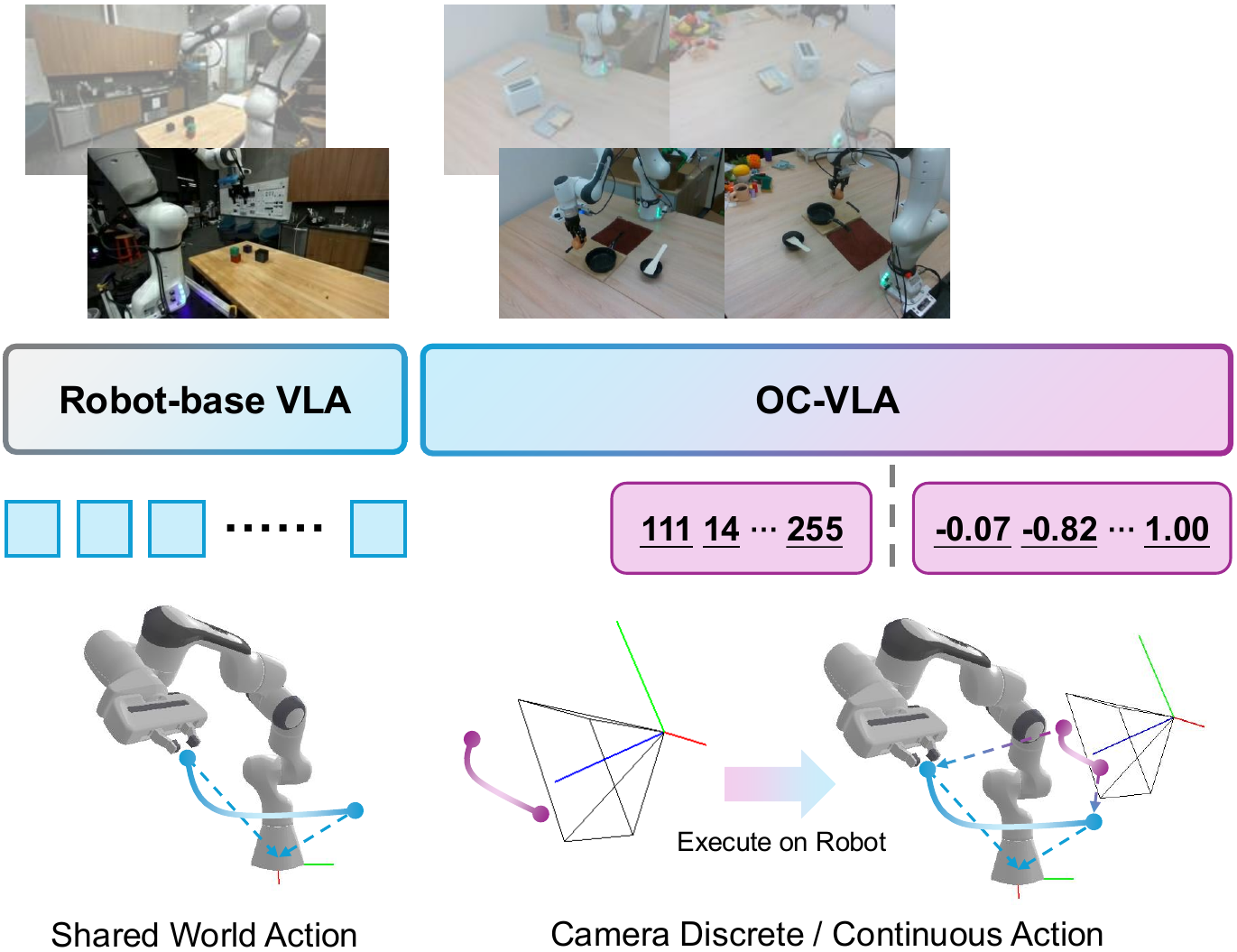} 
\caption{OC-VLA transforms the end effector pose whether defined in a discrete or continuous action space from the robot base coordinate to the third-person camera coordinate, unifying the observation and prediction targets across viewpoints, effectively replacing the usage of shared world action as prediction targets.}
\label{fig02}
\end{figure}
\subsection{Observation-Centric Action Prediction}

In current robotic datasets, action/pose annotations are often defined at a low level, either as joint commands or end-effector poses within the robot base coordinate frame. While these representations are widely used as supervision signals for Vision-Language-Action (VLA) models, they are tightly coupled with specific robot embodiment configurations, \textit{rather than being derived from the observation space}. Consequently, it is difficult for the model to achieve a reasonable projection from image observation to corresponding actions, and thus the model generalization is limited, especially for novel camera views with a large variance from the seen camera views in the training set.

To ground actions in the observation space, it is necessary to first transform the actions from the robot (world) coordinate system into the camera coordinate system. We utilize the extrinsics of the camera to conduct the transformation. Specifically, given two nearby end-effector poses in the world coordinate frame, denoted as  $\mathbf{P}_{{\text{world}}_1}\in R^{4\times4}$  and $\mathbf{P}_{{\text{world}}_2}\in R^{4\times4}$, where the matrix can be converted from a 3D rotation and a translation, the corresponding action can be derived accordingly,

\begin{equation}
\label{eq:world_action}
    \mathbf{A}_{\text{world}} = \mathbf{P}_{{\text{world}}_2} \mathbf{P}_{{\text{world}}_1}^{-1}
\end{equation}

Meanwhile, we can get the corresponding poses in the camera coordinate as follows,

\begin{equation}
\label{eq:translate}
    \mathbf{P}_{{\text{cam}}_2} = \mathbf{T} \mathbf{P}_{{\text{world}}_2}, \mathbf{P}_{{\text{cam}}_1} = \mathbf{T} \mathbf{P}_{{\text{world}}_1}
\end{equation}

where $\mathbf{T}\in R^{4\times4}$ represents the world-to-camera transformation matrix, consisting of a 3D rotation and a translation. $\mathbf{P}$ represents the corresponding matrix. Then, we can obtain the corresponding actions in the camera space.

\begin{equation}
\label{eq:cam_action}
    \mathbf{A}_{\text{cam}} = \mathbf{P}_{{\text{cam}}_2} \mathbf{P}_{{\text{cam}}_1}^{-1}
\end{equation}

Lastly, we convert $\mathbf{A}_{\text{cam}}\in R^{4\times4}$ into the 7-dim actions $\langle \text{x}, \text{y}, \text{z}, \text{roll}, \text{pitch}, \text{yaw}, \text{gripper} \rangle$ for model optimization, where $\text{gripper}$ is for the gripper position. Different from previous end-effector action prediction, the predicted action in our method is in the camera space.

During inference, we transform the actions in the camera space to robot coordinate space for robot control based on the camera calibration.

\subsection{Analysis from Optimization Perspective}
\begin{figure}[h]
\centering
\includegraphics[width=0.98\columnwidth]{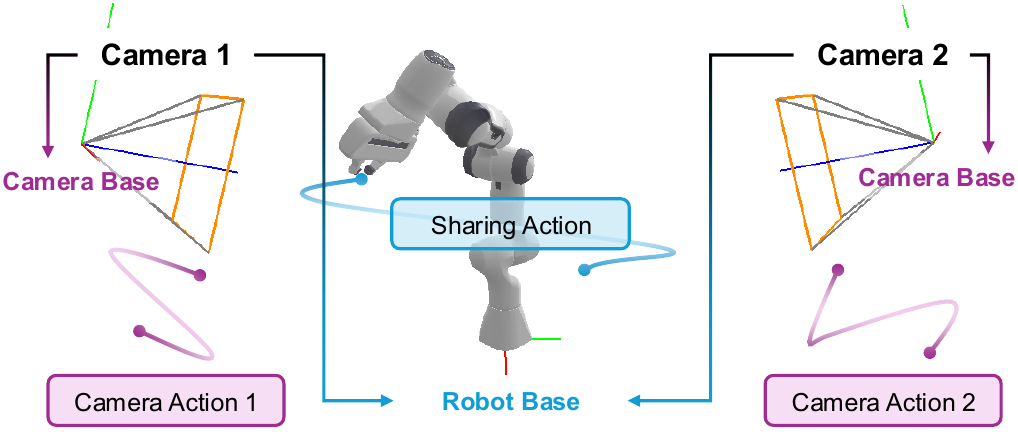} 
\caption{Action translation between robot base coordinate and camera base coordinate. During training, actions are transformed from the robot base coordinate to the camera coordinate and used as groundtruth. During inference, the predicted actions are transformed back from the camera base coordinate to the robot base coordinate for execution on the real robot.}
\end{figure}
In this section, we provide a detailed analysis of the advantages of camera-centric action prediction. In details, we can get $\mathbf{A}_{\text{cam}}$ from equations ~\ref{eq:world_action}, ~\ref{eq:translate} and ~\ref{eq:cam_action} as follow,

\begin{equation}
\label{eq:action}
    \mathbf{A}_{\text{cam}} = \mathbf{T} \mathbf{A}_{\text{world}} \mathbf{T}^{-1}
\end{equation}

where $\mathbf{A}_{\text{cam}}$ is the camera-based action, $\mathbf{A}_{\text{world}}$ is the robot-based action, and $\mathbf{T}$ is the camera world-to-camera transformation matrix.

Meanwhile, given an end-effector pose $\mathbf{P}_{\text{world}}$ of the robot, we can get,

\begin{equation}
\label{eq:pose}
 \mathbf{P}_{\text{cam}} = \mathbf{T} \mathbf{P}_{\text{world}}
\end{equation}

Equations~\ref{eq:action} and ~\ref{eq:pose} present that both the end effector pose and action in world space require the camera transformation matrix $\mathbf{T}$ to be driven from representations in observation space.

In particular, the transformation matrix $\mathbf{T}$ varies across different robot setups. For instance, Droid~\cite{khazatsky2024droid} features 1417 distinct camera viewpoints, requiring the model to internally infer the correct transformation $\mathbf{T}$ for each view to predict actions accurately in the robot's coordinate frame.

Besides, the traditional perception task is based on UV coordinates (image coordinates). According to the intrinsics of the camera, we can obtain the UV coordinate from $(X_{\text{cam}}, Y_{\text{cam}}, Z_{\text{cam}})$. Given that the intrinsic matrix $\mathbf{K}$, the image coordinates $(u, v)$ can be calculated as:

\begin{equation}
u = \frac{f_x \cdot X_{\text{cam}}}{Z_{\text{cam}}} + c_x
\end{equation}

\begin{equation}
v = \frac{f_y \cdot Y_{\text{cam}}}{Z_{\text{cam}}} + c_y
\end{equation}

Where $f_x, f_y$ are the focal lengths in the x and y directions, $c_x, c_y$ are the principal point coordinates (usually the image center). We observe that the camera coordinate can be directly derived from the UV coordinate, and the intrinsic parameters are usually consistent across cameras of the same model. However, translating a point from the camera coordinate system to the robot base coordinate requires the corresponding rotation matrix, which varies with different camera placements. As a result, learning this translation for robot space action prediction becomes more challenging due to the diversity in camera poses. \textit{ In contrast, observation-centric action prediction inherently avoids these issues, offering a more consistent mapping between observation and action.}

\section{Experiments}

In this section, we first provide a detailed description of the pretraining data, followed by an overview of the model architecture for different action spaces. Next, we present the optimization process. Lastly, we present a comprehensive evaluation of the performance of our proposed method on both simulated benchmarks and real-world robotic platforms.
\begin{figure}[h]
\centering
\includegraphics[width=0.98\columnwidth]{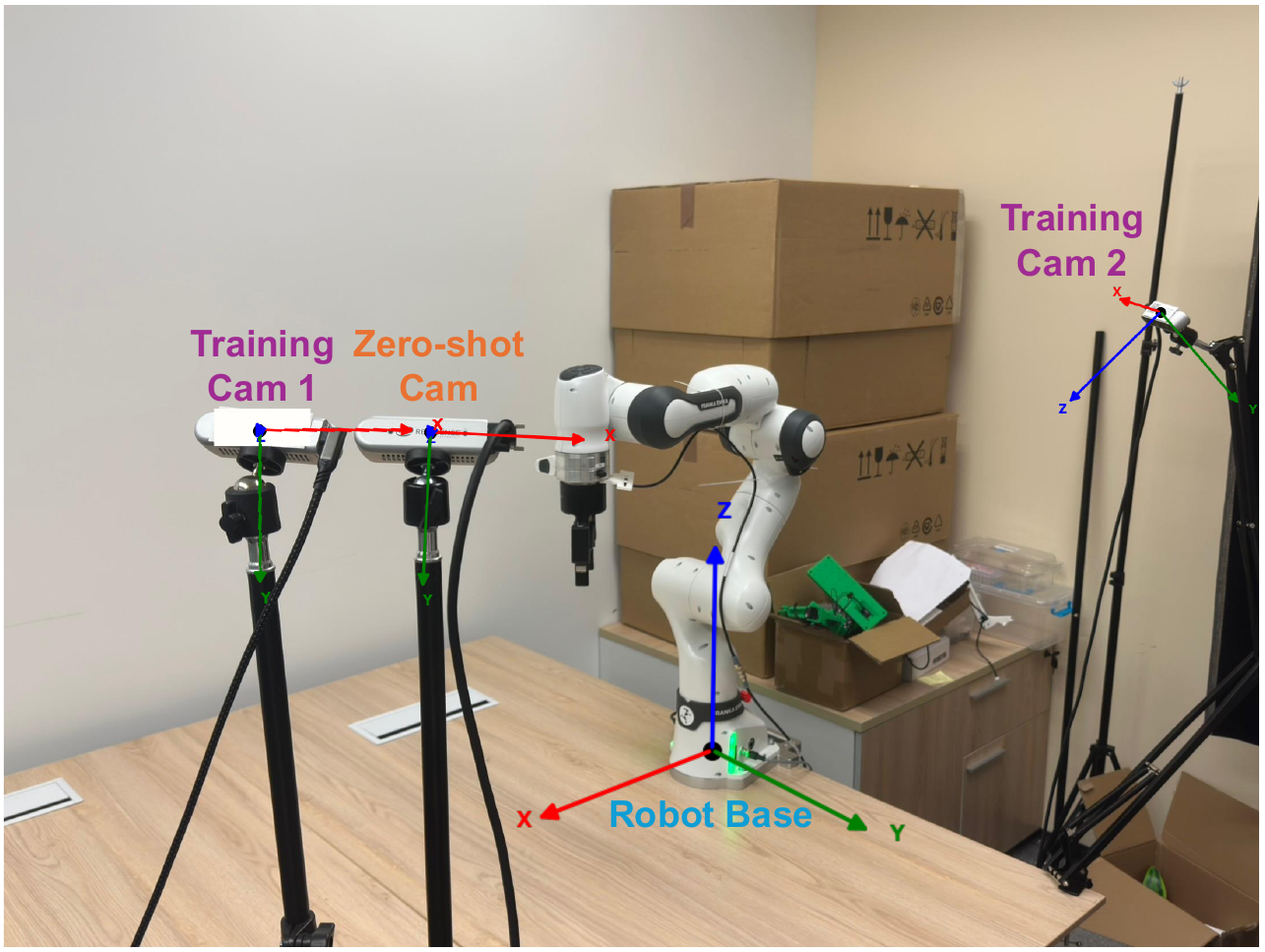} 
\caption{The real-world robot platform with a Franka Emika Panda robot, a Robotiq 2F-85 gripper and multiple RealSense D435i RGB-D cameras.}
\label{fig:setup}
\end{figure}
\subsection{Pretraining Data}
To ensure a comprehensive and fair evaluation of our proposed approach, we incorporate a pretraining stage in selected experiments. Pretraining provides the model with a stronger initialization, which is particularly beneficial when handling complex multimodal inputs and diverse visual contexts. Since our method operates from a third-person perspective and explicitly requires camera extrinsics to transform robot-centric actions into the camera frame, it is crucial to select a dataset that includes such calibration information.

For this purpose, we choose the Droid dataset \cite{khazatsky2024droid} for pretraining. This dataset consists of robotic manipulation trajectories captured from 1417 distinct third-person camera viewpoints, along with their corresponding extrinsic parameters, offering a wide range of visual perspectives and motion patterns. This diversity makes it an ideal choice for evaluating the generalizability and robustness of our observation-centric action prediction framework. Unless otherwise noted, all experiments involving pretrained models are initialized using weights obtained through pretraining on the Droid dataset~\cite{khazatsky2024droid}.
\subsection{Model Details}

In our experiments, we employ a typical lightweight VLM architecture, with distinct designs for the continuous and discrete action spaces. In the following, we detail the model implementations for each action space.

For the continuous action space model, we adopt a diffusion policy. In addition to language and image tokens, we concatenate the current timestep and the noise-perturbed action as inputs to the causal transformer. The entire transformer functions as a Diffusion Transformer (DiT) \cite{peebles2023scalable}, which iteratively denoises the input over multiple steps to generate the final end-effector action.

For the discrete action space model, we pad zero vectors to align the action size after processing language and image inputs. The combined sequence is then fed into the Transformer. Despite using a causal mask during training, the model predicts the entire action sequence in a single pass, rather than autoregressively. This design enhances both semantic consistency across tokens and computational efficiency.

\subsection{Optimization Details}

The training objectives vary depending on the type of action space used. For models with a continuous action space, the objective is to minimize the mean squared error (MSE) between the robot's action (augmented with standard Gaussian noise) and the predicted noise, using DDPM \cite{ddpm} with 100 timesteps. In contrast, for models with a discrete action space, the robot actions are normalized to a predefined range and quantized into discrete bins. The objective here is to minimize the cross-entropy loss between the predicted discrete actions and the ground-truth labels.

For diffusion evaluation, we use DDIM \cite{ddim} with 10 timesteps during inference. The model is optimized using AdamW \cite{adamw} for 30,000 steps, with learning rates of $1e-4$ for both the causal Transformer and Q-Former, and $1e-5$ for DINOv2. Training is conducted with a batch size of 2048 across 8 NVIDIA A100 GPUs, with 256 samples per GPU. The model predicts actions in the third-person camera base coordinate, while the baseline model predicts actions in the robot base coordinate.
\subsection{Simulation Evaluation}
\begin{table*}[h]
\caption{Comparison on ManiSkill2 under Success rate. SingleYCB indicates PickSingleYCB, ClutterYCB indicates PickClutterYCB, SingleEGAD indicates PickSingleEGAD. Coord indicates the selected coordinate while training. Continuous indicates whether the action prediction is continuous or discrete.}

\centering
\begin{tabular}{c|c|cccccc}
{\bf Coord} & {\bf Continuous}  & {\bf All} & { \bf PickC} & {\bf StackC} & {\bf SingleYCB} & {\bf ClutterYCB} & {\bf EGAD}\\
\midrule
Robot & \checkmark &   45.2\%  &  71.0\%& 62.0\%  & 30.0\%   & 15.0\%  &  48.0\%  \\
Camera & \checkmark  &  {\bf 53.2\%} &  {\bf 88.0\%} &  {\bf 65.0\%} &  {\bf 46.0\% }&  {\bf 19.0\%} &  48.0\%    \\
\hline
Robot &  \texttimes &  38.6\%  &  61.0\% & 51.0\%  & 28.0\%  & 8.0\%  & 45.0\%  \\
Camera &  \texttimes & {\bf 52.4\%} & {\bf 80.0\%}  &  {\bf 65.0\%} & {\bf 48.0\%}  & {\bf 19.0\%}  & {\bf 50.0\%}  \\



\end{tabular}
\label{tab:maniskill_ab}
\end{table*}
\subsubsection{Simulation Dataset}
For simulated evaluation, we select ManiSkill2 \cite{gu2023maniskill2} to assess the effectiveness and generalization capabilities of our proposed approach. ManiSkill2, the successor to the original SAPIEN ManiSkill \cite{mu2021maniskill} benchmark, has become a widely recognized and authoritative platform for evaluating the generalization performance of embodied agents in robotic manipulation. Meanwhile, ManiSkill2 includes 20 diverse task families, covering a broad range of real-world manipulation scenarios. Additionally, ManiSkill2 supports rendering observations from randomly sampled camera viewpoints, making it a suitable choice for our evaluation.
\subsubsection{Setup}
To construct our benchmark, we select five representative tasks from the ManiSkill2 suite: PickCube-v0, StackCube-v0, PickSingleYCB-v0, PickClutterYCB-v0, and PickSingleEGAD-v0. We generate a pool of 300,000 randomly configured third-person camera viewpoints. For each trajectory, 20 cameras are randomly sampled to render the demonstration, resulting in a dataset comprising over 40,000 unique trajectories. We partition the generated data into training and validation sets using a 19:1 ratio. Care is taken to ensure that each task family is represented in both sets, and that trajectories rendered from different camera viewpoints are distributed across the splits, thereby preventing data leakage. To address data imbalance, we replicate trajectories from underrepresented task families to equalize the number of samples across tasks during training. For closed-loop evaluation, we randomly sample 100 trajectories from the validation set for each task family, resulting in an evaluation set of 500 trajectories. This evaluation benchmark is used to measure the success rate of the model across different manipulation tasks.
\subsubsection{Comparison}
Given the domain gap between Droid and Maniskill2, both the continuous and discrete action space models are trained from scratch in this evaluation. We conduct a comparative analysis of their performance under two supervision regimes: one using robot actions defined in the robot base coordinate frame, and the other using robot actions transformed into the third-person camera coordinate frame as the prediction targets. Table~\ref{tab:maniskill_ab} shows the performance of the comparison of the different models with different prediction target. The results demonstrate that, regardless of the type of action space used, employing robot actions defined in the third-person camera coordinate frame as prediction targets consistently improves task success rates. \textit{ This improvement is particularly pronounced in models utilizing a discrete action space, where we observe an increase in success rate of about 14\%}.
\subsection{Real Robot Evaluation}
\begin{table*}[h]
\caption{Quantitative results in Real robot experiments. Methods annotated with "(var)" indicate results obtained under zero-shot camera evaluation, while those without the annotation correspond to evaluations conducted using the Training Cam 1. Robot Base and Camera Base indicates the model we built in robot base coordinates and third-person camera base coordinate following Dita~\cite{hou2025dita}, respectively. The mapping between task IDs and their corresponding simple task descriptions is as follows. Task 1: Pick up the carrot into the box. Task 2: Open box. Task 3: Close box. Task 4: Press button. Task 5: Pick up the cup into compartment. Task 6: Cook. Task 7: Bake the bread. Task 8: Push the book. Task 9: Put the marker and insert into pen box. Task 10: Stack dolls. Task 11: Stack bowls. Task 12: Push the toy car. Task 13: Pour the water. Task 14: Fold the towel. Task 15: Use the microwave. A more detailed description of each task is provided in the appendix.}
\begin{center}
\begin{tabular}{l|c|ccccccc}

{\bf Method} & {\bf Avg}  & {\bf Task 1} & {\bf Task 2} & {\bf Task 3}& {\bf Task 4} & {\bf Task 5} & {\bf Task 6} & {\bf Task 7} \\
\midrule
{OpenVLA-OFT} & 63.3\% &  100.0\%    & 80.0\%  & 90.0\%  & 80.0\%  & 80.0\% &  80.0\% & 60.0\% \\
{OpenVLA-OFT (var)} & 42.0\% &  90.0\%    & 70.0\%  & 60.0\%  & 40.0\%  & 50.0\% &  50.0\% & 10.0\% \\
{$\pi_0$} & 50.7\% &   50.0\%   & 70.0\%  & 80.0\%  & 60.0\% & 60.0\% &  70.0\% & 80.0\% \\
{$\pi_0$ (var)} & 34.7\% &   20.0\%   & 40.0\%  & 60.0\%  & 40.0\% & 30.0\% &  30.0\% & 60.0\% \\
\midrule
{Robot Base} & 58.0\% &  70.0\% & 70.0\%  &  90.0\% & 60.0\% & 60.0\% & 60.0\% & 60.0\% \\
{Robot Base (var)} & 41.3\% & 40.0\%  & 50.0\%  & 70.0\%  &  60.0\% &  40.0\% & 60.0\% & 60.0\% \\
{Camera Base (OC-VLA, ours)}& {\bf 68.0\%} &   80.0\%   & 80.0\%  & 100.0\%  & 80.0\%  &  80.0\% &  70.0\% & 60.0\% \\
{Camera Base (var)} & 54.0\% & 70.0\%  & 70.0\%  & 100.0\%  & 70.0\%  & 60.0\% & 60.0\% & 70.0\%  \\
 \multicolumn{2}{}{}    & & &  &  &  &  &   \\ 
  \multicolumn{1}{l}{\bf Method} & \multicolumn{1}{c}{\bf Task 8}  & {\bf Task 9} & {\bf Task 10} & {\bf Task 11}& {\bf Task 12} & {\bf Task 13} & {\bf Task 14} & {\bf Task 15} \\
  \midrule
  \multicolumn{1}{l}{OpenVLA-OFT} & \multicolumn{1}{c}{70.0\%} &  50.0\%    &  20.0\% & 20.0\% &80.0\%  & 50.0\%  & 60.0\% & 30.0\% \\
 \multicolumn{1}{l}{OpenVLA-OFT  (var)} & \multicolumn{1}{c}{40.0\%} &  50.0\%    &  10.0\% & 10.0\% &30.0\%  & 50.0\%  & 50.0\% & 20.0\% \\
\multicolumn{1}{l}{$\pi_0$} & \multicolumn{1}{c}{60.0\%} &  40.0\%  & 10.0\%  & 20.0\%  & 40.0\% & 50.0\% & 60.0\% & 10.0\% \\
\multicolumn{1}{l}{$\pi_0$ (var)} & \multicolumn{1}{c}{60.0\%} &  30.0\%  & 10.0\%  & 10.0\%  & 50.0\% & 40.0\% & 30.0\% & 10.0\% \\
\midrule
\multicolumn{1}{l}{Robot Base} & \multicolumn{1}{c}{60.0\%} & 60.0\% & 20.0\%  & 40.0\% & 50.0\% & 60.0\% & 90.0\% & 20.0\%  \\
\multicolumn{1}{l}{Robot Base (var)} & \multicolumn{1}{c}{30.0\%} & 30.0\%  & 10.0\%  &  20.0\% & 30.0\%  & 50.0\%  & 50.0\% & 20.0\% \\
\multicolumn{1}{l}{Camera Base (OC-VLA, ours)} & \multicolumn{1}{c}{70.0\%} & 60.0\% &  40.0\% & 50.0\%  & 70.0\%  & 60.0\% &  90.0\% & 30.0\% \\
\multicolumn{1}{l}{Camera Base (var)} & \multicolumn{1}{c}{40.0\%} & 50.0\%  & 20.0\%  &  30.0\% & 20.0\%  & 60.0\% & 70.0\% & 20.0\% \\
\end{tabular}
\end{center}
\label{tab:RealRobot1}
\end{table*}

\begin{table*}[h]
\caption{Real robot experiments of different camera views. Fixed Camera means no camera perturbations while data collection. The meanings of the methods and the Task ID mappings follow the same convention as in Table \ref{tab:RealRobot1}.}
\begin{center}
\begin{tabular}{l|c|cccccccc}

{\bf Method}  & {\bf Avg} & {\bf Task1} & {\bf Task2} & {\bf Task3}& {\bf Task4} & {\bf Task5} & {\bf Task6}& {\bf Task7} & {\bf Task8} \\ 
\midrule
Robot Base(Fixed Camera, From Table \ref{tab:RealRobot1})  & 66.3\%  & 70.0\%  & 70.0\%  & 90.0\%  & 60.0\% & 60.0\% & 60.0\% & 60.0\% & 60.0\% \\
Cam Base(Fixed Camera, From Table \ref{tab:RealRobot1})  & 77.5\%  & 80.0\%  & 80.0\%  & 100.0\%  & 80.0\% & 80.0\% & 70.0\% & 60.0\% & 70.0\%  \\
\midrule
Robot Base(Camera Perturbations)  & 61.3\%  & 80.0\%  & 70.0\%  & 50.0\%  & 70.0\% & 40.0\% & 60.0\% & 50.0\% & 70.0\% \\
Cam Base(Camera Perturbations)  & 73.8\%  & 80.0\%  & 80.0\%  & 70.0\%  & 90.0\% & 80.0\% & 60.0\% & 60.0\% & 70.0\%  \\
\end{tabular}
\end{center}
\label{tab:RealRobot3}
\end{table*}
\subsubsection{Setup}

We evaluate OC-VLA on a real-world Franka Robot setup, which comprises a 7-DoF tabletop Franka Emika Panda robot arm equipped with a Robotiq 2F-85 gripper as shown in Figure~\ref{fig:setup}. Three RealSense D435i RGB-D cameras are positioned to capture the task environment from multiple third-person perspectives. Specifically, two cameras are used for both data collection and few-shot evaluation, while the remaining camera is reserved exclusively for zero-shot evaluation.
\subsubsection{Data Collection and Model Finetuning}
We adopt a demonstration-based approach to collect two datasets from different viewpoints using Training Camera 1 and Training Camera 2, respectively. For the dataset collected with Camera 1, we record trajectories for 15 distinct tasks while keeping the camera position fixed throughout the entire data collection process. In contrast, the dataset collected with Camera 2 consists of trajectories for 8 tasks, during which we introduce slight perturbations to the camera position to simulate minor viewpoint variations. The collected tasks span a diverse set of categories, including pick \& place, pouring, stacking, pick \& rotation, pull \& push, as well as other long-horizon tasks, aiming to comprehensively evaluate the true performance of the model. A detailed list of tasks is provided in the appendix. Following Dita~\cite{hou2025dita}, for each task in both datasets, we collect 10 demonstration trajectories, aiming to evaluate the model fitting ability under a 10-shot setting.

For model finetuning, we fine-tune the model pretrained on the Droid dataset, using either end effector actions defined in the third-person camera coordinate or those in the robot base coordinate as prediction targets. Both models are optimized with AdamW \cite{adamw} for 20,000 steps with a batch size of 512. For a fair performance comparison, we also fine-tune the pretrained versions of OpenVLA-OFT \cite{kim2024openvla}, $\pi_{0}$ \cite{black2024pi_0} on our collected datasets, using their official training protocols. These models serve as baselines in our evaluation.
\begin{figure}[h]
\centering
\includegraphics[width=0.98\columnwidth]{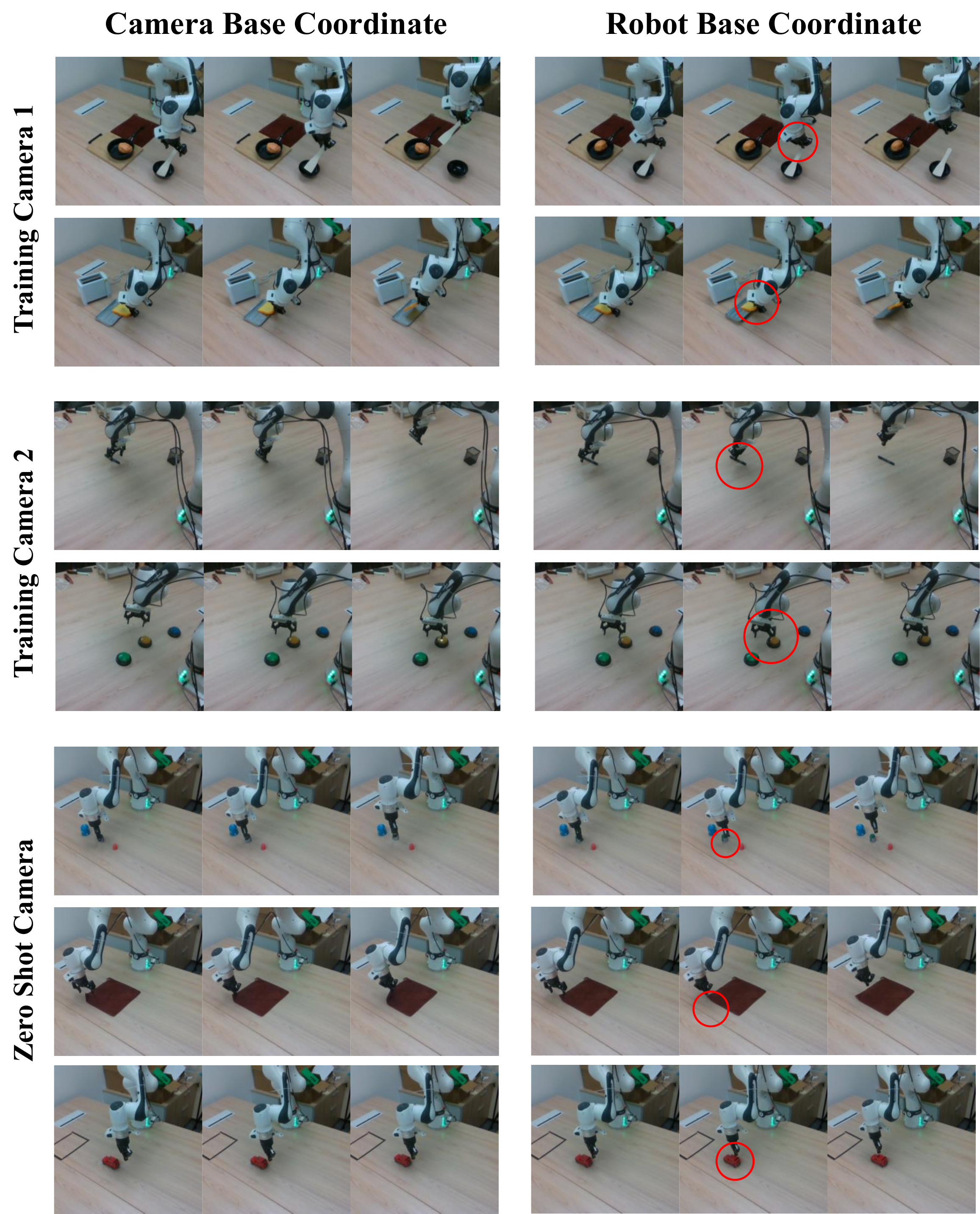} 
\caption{A qualitative comparison in real-robot experiments. Failures are highlighted with red circles. }
\label{fig:qualitative}
\end{figure}
\subsubsection{Quantitative Evaluation and Comparison}
The evaluations are organized into the following three main settings:
\begin{itemize}
\item { \bf Fixed Camera Viewpoint.} We fine-tune all models using 15 task demonstrations collected from Camera 1 and perform a unified evaluation. For each task, we conduct 10 trials and measure performance by computing the task success rate. In this setting, the camera viewpoint remains fixed and identical throughout both the fine-tuning and evaluation phases.

\item {\bf Slight Camera Perturbations} To further validate the robustness of our method, we introduce slight variations in the camera viewpoints. Specifically, we fine-tune the models using 8 task demonstrations collected from Camera 2, each exhibiting minor differences in camera placement. For evaluation, we position the camera in a similar configuration to the fine-tuning setup and recalibrate the camera to obtain updated extrinsic parameters. The camera remains fixed throughout the evaluation process.

\item {\bf Novel Camera Viewpoint.} To assess the model’s robustness to changes in camera perspective, we conduct zero-shot evaluations using models fine-tuned with demonstrations from Camera 1. As illustrated in the Figure \ref{fig:setup}, we introduce a novel, previously unseen camera mounted near Camera 1, and perform all evaluations under this new fixed viewpoint without any additional fine-tuning.
\end{itemize}
{\bf Fix Camera View.} As shown in the Table~\ref{tab:RealRobot1}, under the 10-shot setting with a fixed camera viewpoint, the model fine-tuned using robot base coordinate actions already demonstrates competitive performance. However, when the prediction target is switched from robot-base coordinate actions to camera-base coordinate actions, the model achieves a further 10\% improvement in the metric of success rate, surpassing the best-performing baseline, OpenVLA-OFT, fine-tuned on the same data. This indicates that our method can partially compensate for the limited pretraining data and model size by improving data efficiency.

{\bf Novel Camera View.} For novel camera view, all models exhibit varying degrees of performance degradation in Table~\ref{tab:RealRobot1}, as expected. Notably, OpenVLA-OFT, which performs as the best baseline under the 10-shot setting, suffers a performance drop of over 20\%. In contrast, our method shows only a 14\% decrease, outperforming all baselines in this setting. \textit{ These results highlight the added robustness to camera viewpoint changes when the model is trained to predict actions in the camera base coordinate frame}.

{\bf Camera Perturbations.} Furthermore, the results in Table \ref{tab:RealRobot3} demonstrate the advantage of using camera-base coordinate actions as prediction targets when there is variance in camera viewpoints within the fine-tuning data. Although the overall performance is slightly lower than that under strictly fixed-view conditions, the relative benefit of camera-based supervision increases, underscoring the generalizability of our approach in more realistic and variable settings.
\subsubsection{Qualitative Comparison}
Figure \ref{fig:qualitative} shows a comparison between OC-VLA and the baseline method under the robot base coordinate across different evaluation conditions and camera viewpoints. The results illustrate that OC-VLA offers improved robustness for fine-grained manipulation under a variety of settings. While baseline methods often fail to successfully complete tasks due to inaccurate grasp localization—especially under camera perturbations, OC-VLA consistently identifies more precise grasp positions. This advantage is particularly evident when there is variance in camera viewpoints: whereas baseline models begin to exhibit subtle errors, OC-VLA remains resilient and is able to complete the task successfully.

\section{Conclusion}
In this paper, we propose Observation-Centric VLA (OC-VLA), a simple yet effective framework that grounds action predictions in the camera base coordinate, addressing the spatial misalignment between perception and action in existing VLA models. OC-VLA introduces no architectural overhead and integrates seamlessly with existing pipelines. Extensive experiments show that OC-VLA significantly improves the cross-view generalization and enhance robustness under viewpoint shifts, showing the practical utility of OC-VLA and its strong potential for generalist robot policies.


\bibliographystyle{IEEEtran}
\bibliography{reference}
\clearpage
\section*{APPENDIX}

\subsection{Model Structure and Details}
\begin{figure*}[!h]
\centering
\includegraphics[width=0.98\textwidth]{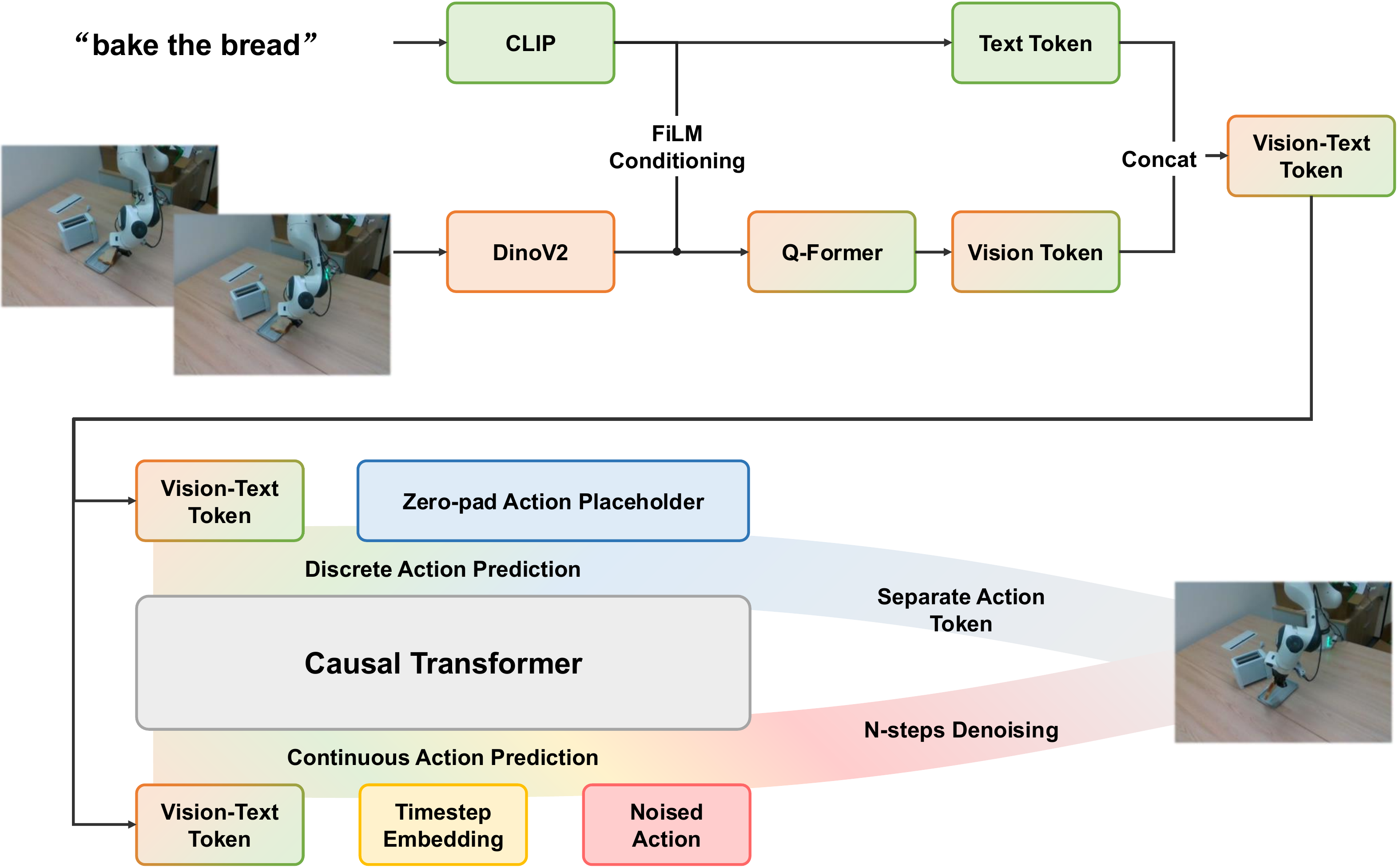} 
\caption{Model Structure. We use the same model structure which is followed Dita to evaluate our method on both continuous action space and discrete action space, but treat the Transformer as different function module. The Transformer is served as a Diffusion Transformer (DiT) in the model predicting the actions in the continuous action space but as a non-autoregressive Transformer in which working in the discrete action space.}
\label{modelstructure}
\end{figure*}
In this section, we provide a detailed description of the model architecture and implementation used in our experiments. The overall structure follows the design of Dita without modification~\cite{hou2025dita}, as illustrated in Figure \ref{modelstructure}. Specifically, the model takes as input only a language description and an RGB image from a third-person camera.

The language description is encoded using a pretrained CLIP text encoder~\cite{radford2021learning}, while the RGB image is first resized to 224×224 and then processed through a pretrained DINOv2~\cite{oquab2023dinov2} vision encoder. The resulting image features are passed into a 4-layer Q-Former~\cite{li2023blip}, which serves to reduce the number of image tokens and control the overall model size. The number of image tokens is reduced to 32. The Q-Former is trained from scratch. Additionally,  a FiLM layer~\cite{perez2018film} is injected into each Q-Former block, where the encoded language embedding is used as conditioning input to guide the selection and compression of visual features. The processed language and image features are then jointly fed into a LLaMA2-style Transformer~\cite{touvron2023llama}, which produces the final predicted action for execution by the robot. This Transformer consists of 12 layers with a hidden size of 768, and operates under a causal masking scheme. The total model size is approximately 334M parameters, with only the CLIP text encoder frozen during training.

To ensure fair comparison, we adopt this unified architecture for both the continuous and discrete action space experiments, with only minor differences in how the Transformer component is used. For continuous action space models, we follow Dita and treat the Transformer as a Diffusion Transformer (DiT). During training, the ground-truth action is perturbed with noise using a 100-timestep DDPM scheduler~\cite{ddpm}, and then input into the DiT along with the timestep embedding and the preprocessed language and image features. The DiT is trained to predict the added noise. During inference, we use a 10-timestep DDIM scheduler~\cite{ddim} for efficient denoising, which previous work has shown to maintain strong performance with reduced computational cost. For discrete action space models, the Transformer is not autoregressive, predicting all action tokens in a single forward pass. The action is first normalized into a fixed range and tokenized. This formulation ensures efficient inference and explicitly decouples dependencies between individual action tokens.

\subsection{Full Pipeline of Our method}
\begin{figure*}[h]
\centering
\includegraphics[width=0.98\textwidth]{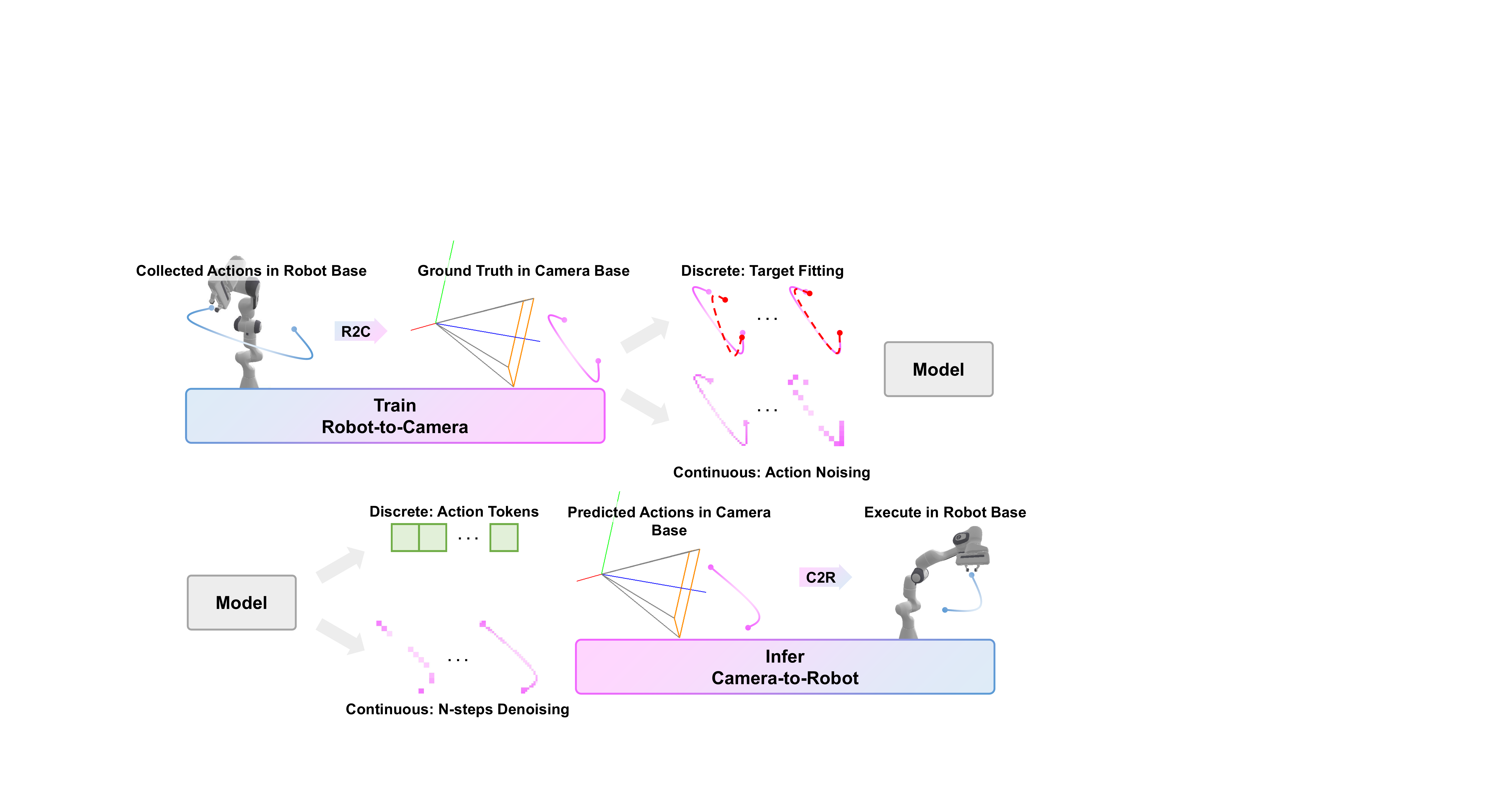} 
\caption{Full Pipeline of our method. We introduce OC-VLA framework, aligning the observation space and the prediction target with the camera extrinsic calibration matrix. It is simple and efficient, improve the performance of the VLA models without any extra GPU consumption.}
\label{pipeline}
\end{figure*}
In this work, we propose the Observation-Centric Vision-Language-Action (OC-VLA) framework, which leverages the extrinsic calibration matrix of a third-person camera to transform the end-effector pose from the robot base coordinate to the camera base coordinate . The transformed pose is then used as the prediction target for the model, thereby aligning the observation space with the action prediction target. The overall architecture of the proposed framework is illustrated in Figure \ref{pipeline}.
Our framework introduces a minor distinction between the training and inference stages, which consists of the following key steps. During training, since the end-effector poses in most robotic datasets are defined in the robot base coordinate, we first apply the extrinsic calibration matrix of the third-person camera to transform the pose into the camera coordinate frame. This transformation process is described in detail in the Method Section. The transformed pose in the camera base coordinate is then used as the groundtruth for supervision, aligning the model's prediction target with the visual observation space. During inference, the model outputs an end-effector pose in the camera base coordinate. However, real-world robotic systems typically require poses expressed in the robot base coordinate. To bridge this gap, we apply a post-processing step that transforms the predicted pose back from the camera base coordinate to the robot base coordinate using the same extrinsic matrix. The converted pose is then sent to the physical robot for execution.

Our approach is simple, efficient, and plug-and-play, requiring no additional GPU overhead and minimal integration effort. It offers strong potential for practical adoption in VLA systems, especially in settings involving diverse or dynamic camera viewpoints.
\subsection{Simulation Benchmark Experiments}
\subsubsection{Dataset Visualization}
\begin{figure*}[!h]
\centering
\includegraphics[width=0.98\textwidth]{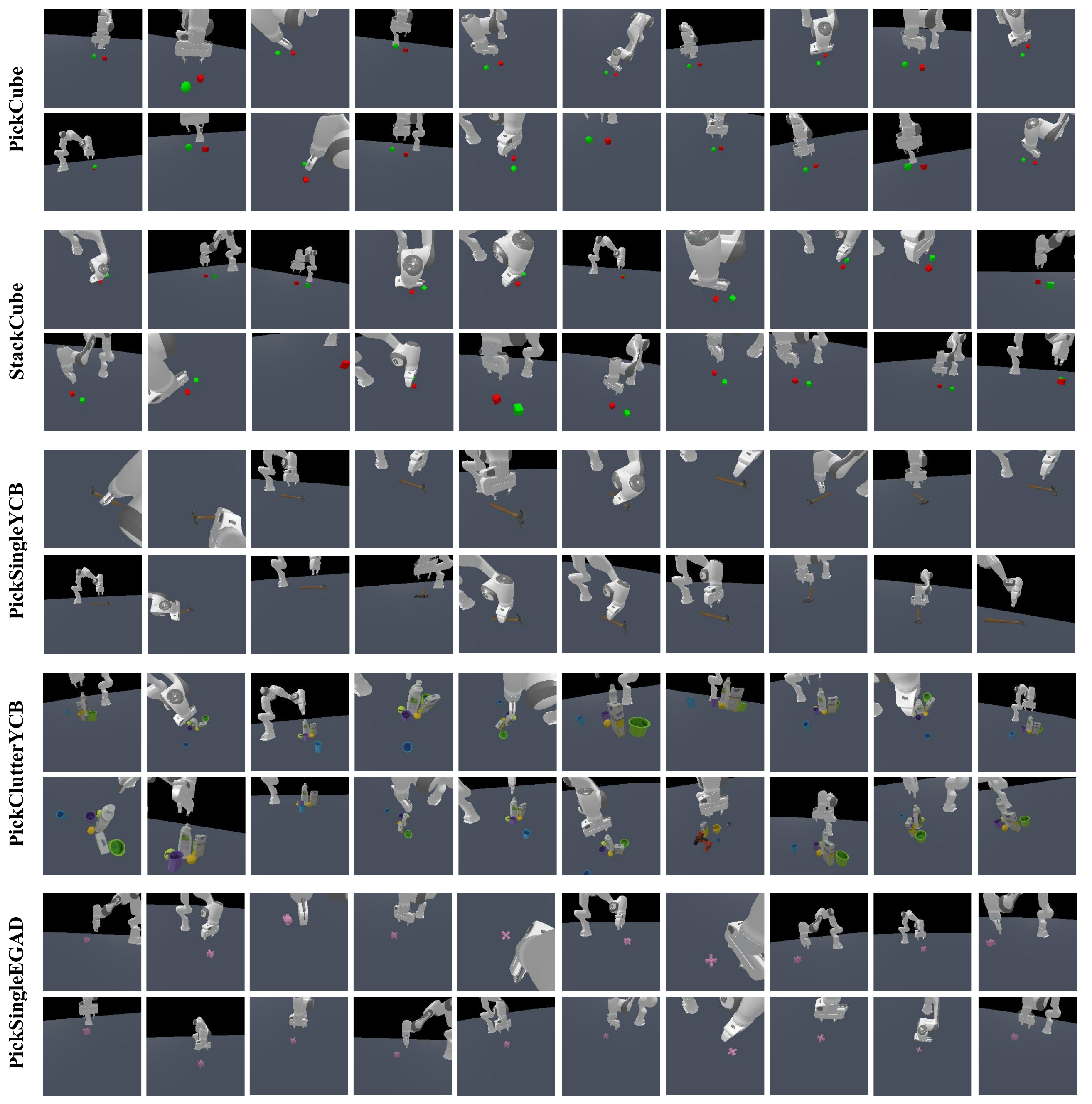} 
\caption{Visualization of the ManiSkill2 Dataset. We generate a third-view camera pool in the Simulated Environment and sample 20 cameras for each of the trajectory to render the data as our dataset.}
\label{ManiVisual}
\end{figure*}
For the simulation experiments, we utilize the ManiSkill2 dataset~\cite{gu2023maniskill2}, a benchmark built on the SAPIEN simulator that supports flexible third-person camera placement and trajectory rendering—making it particularly well-suited for evaluating our proposed method.

Leveraging these features, we construct a new dataset by selecting five task families from ManiSkill2: PickCube, StackCube, PickSingleYCB, PickSingleEGAD, and PickClutterYCB. To introduce sufficient visual diversity, we generate a camera pool containing 300,000 randomly sampled third-person viewpoints, and for each trajectory, we randomly select 20 camera poses from this pool to render demonstrations.

The resulting dataset consists of approximately 40,000 unique trajectories, with 5\% held out as a validation set. Throughout the dataset construction process, we ensure balanced distribution across task families and strict separation between training and validation sets to prevent data leakage. Figure \ref{ManiVisual} provides representative visual examples from our generated dataset.
\subsubsection{Qualitative Comparison}
\begin{figure*}[h]
\centering
\includegraphics[width=0.98\textwidth]{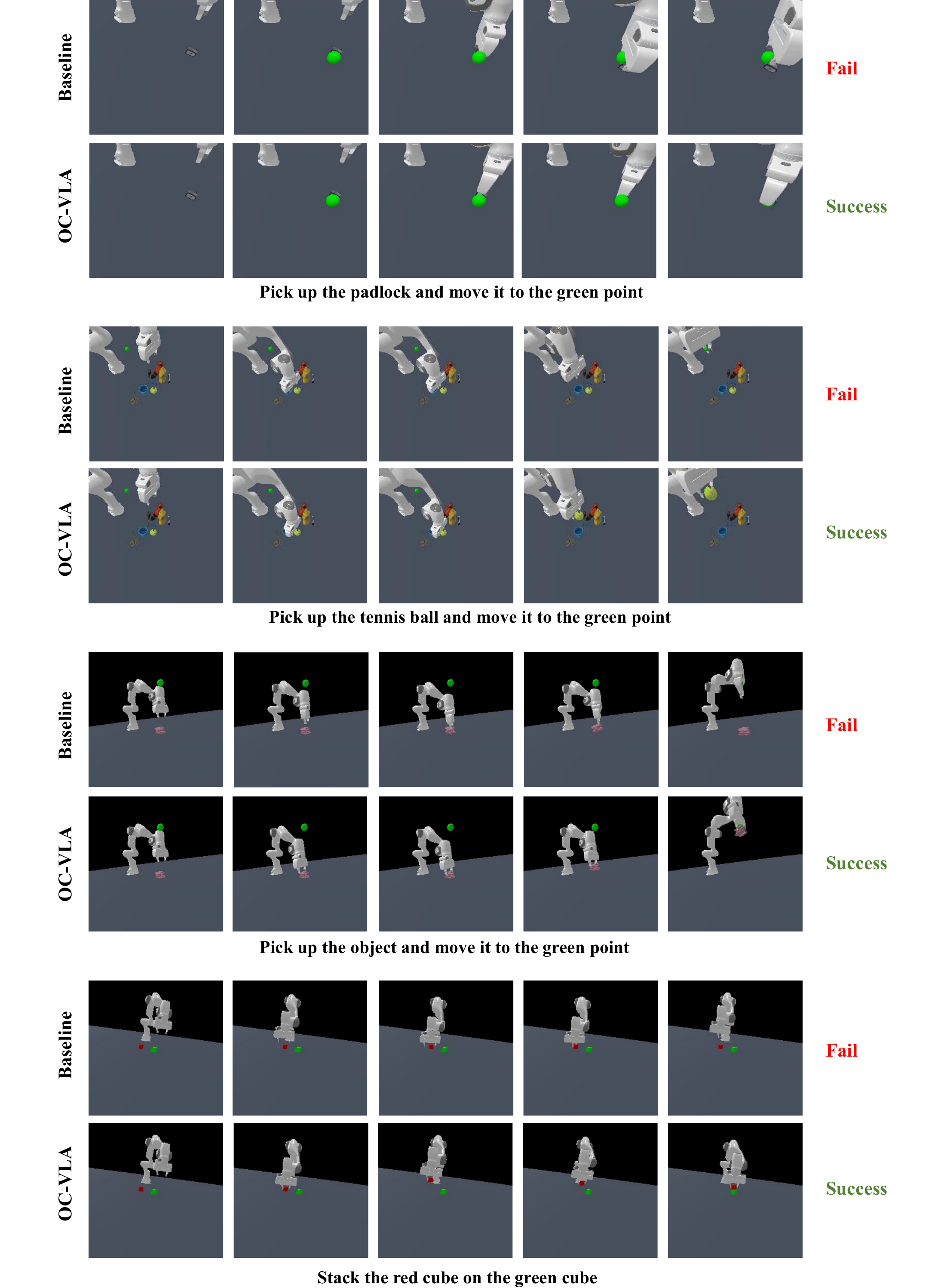} 
\caption{Qualitative Comparison on ManiSkill2 of OC-VLA and Baseline. OC-VLA show better performance on the grasp pose and searching for the goal point.}
\label{ManiQualitative}
\end{figure*}
In Figure \ref{ManiQualitative}, we demonstrate some qualitative results of our method. As illustrated in the figure, the alignment between the observation space and the prediction target enables OC-VLA to produce more accurate grasp poses and better end-effector alignment. The model consistently identifies the correct goal point with higher precision, which contributes to the superior performance observed on this benchmark.
\subsubsection{Ablation Study}

To comprehensively evaluate the performance of OC-VLA under diverse conditions, we conduct a series of ablation studies on the model configured with a discrete action space. The results are summarized in Table \ref{tab:maniskill_ab2}.

Specifically, we investigate the impact of three key factors during training: the observation sequence length, the trajectory length, and whether the ViT encoder is frozen during optimization. Across all settings, OC-VLA, which uses action targets represented in the camera base coordinate, consistently outperforms the baseline model that operates in the robot base coordinate. These results highlight the generalizability and robustness of OC-VLA across different training conditions.
\begin{table*}[h!]
\centering
\caption{Ablation Study on ManiSkill2. SingleYCB indicates PickSingleYCB, ClutterYCB indicates PickClutterYCB, SingleEGAD indicates PickSingleEGAD. Coord indicates the selected coordinate while training. \#Obs, \#Traj, \#Freeze ViT indicates the observation length, the trajectory length and whether freezeing the ViT backbone while training.}
\begin{tabular}{c|c|c|c|cccccc}

\toprule
{\bf Coord}  &{ \bf \#Obs } & { \bf \#Traj} &{\bf \# Freeze ViT} & {\bf All} & { \bf PickC} & {\bf StackC} & {\bf SingleYCB} & {\bf ClutterYCB} & {\bf EGAD}\\
\hline

Robot &  2 & 2 & \texttimes &  38.6\%  &  61.0\% & 51.0\%  & 28.0\%  & 8.0\%  & 45.0\%  \\
Camera & 2 & 2 & \texttimes & {\bf 52.4\%} & {\bf 80.0\%}  &  {\bf 65.0\%} & {\bf 48.0\%}  & {\bf 19.0\%}  & {\bf 50.0\%}  \\

\hline

Robot &  2 & 2 & \checkmark &  16.6\%  &  34.0\% & 24.0\%  &  10.0\% & 6.0\% &  9.0\%\\
Camera &  2 & 2 & \checkmark & {\bf 27.8\%} &  {\bf 51.0\%} & {\bf 49.0\%}  &  {\bf 14.0\%} & {\bf 9.0\%}  &  {\bf 16.0\%} \\

\hline
Robot  &  2 & 16 & \texttimes & 16.4\%   & 23.0\%  & 29.0\%  & 15.0\%  &  1.0\% & 14.0\%  \\
Camera  &  2 & 16 & \texttimes & {\bf 39.4\%} & {\bf 63.0\%}  &  {\bf 58.0\%} & {\bf 27.0\%}  & {\bf 15.0\%}  & {\bf 34.0\%}  \\
\hline
Robot & 3 & 3 & \texttimes  & 33.0\%  & 54.0\% & 32.0\%  & 28.0\%   & 6.0\% & 45.0\%   \\
Camera & 3 & 3 & \texttimes &  {\bf 51.8\%}  & {\bf 77.0\%}  & {\bf 75.0\%}  & {\bf 43.0\%}  & {\bf 9.0\%}  & {\bf 55.0\%}  \\


\end{tabular}
\label{tab:maniskill_ab2}
\end{table*}
\subsection{Real Robot Experiments}

\subsubsection{Tasks Details for Real Franka Arm Evaluation}

\begin{figure*}[h]
    \centering
    \begin{tabular}{c}
\includegraphics[width=1\linewidth]{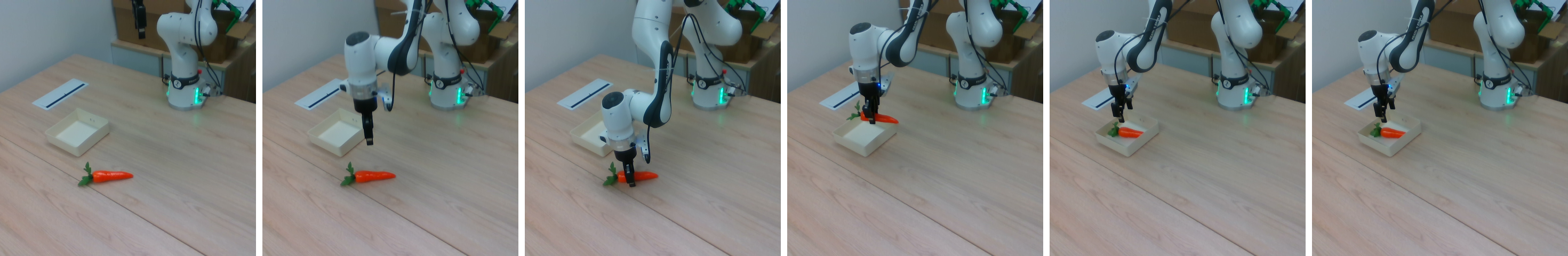}\vspace{1.5mm}\\
\bf Pick up the carrot into the box.
\vspace{2mm}\\

\includegraphics[width=1\linewidth]{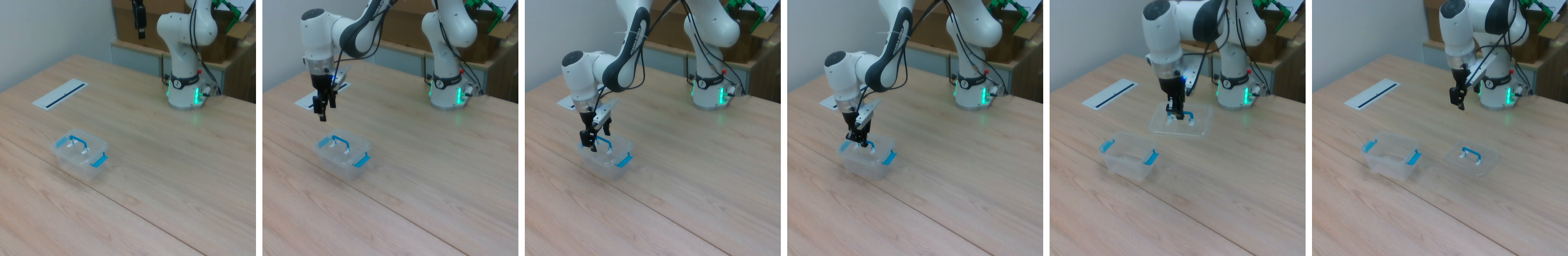}\vspace{1.5mm}\\
\bf Open the storage box.
\vspace{2mm}\\

\includegraphics[width=1\linewidth]{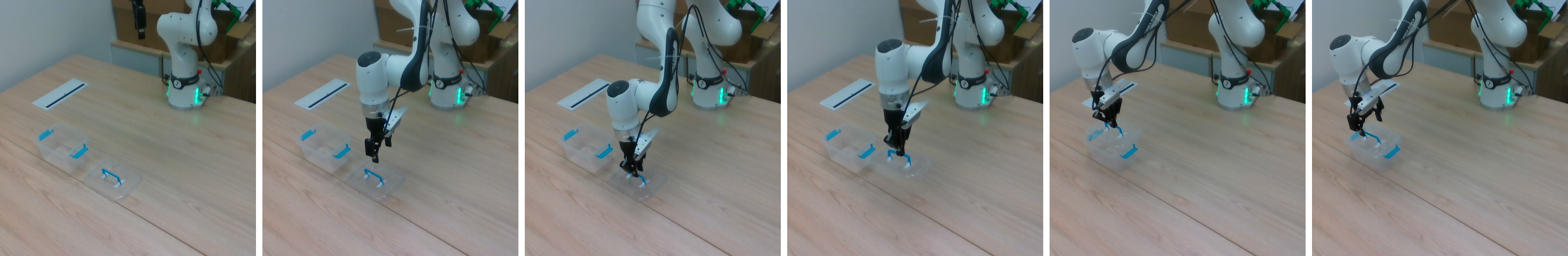}\vspace{1.5mm}\\
\bf Close the storage box.
\vspace{2mm}\\

\includegraphics[width=1\linewidth]{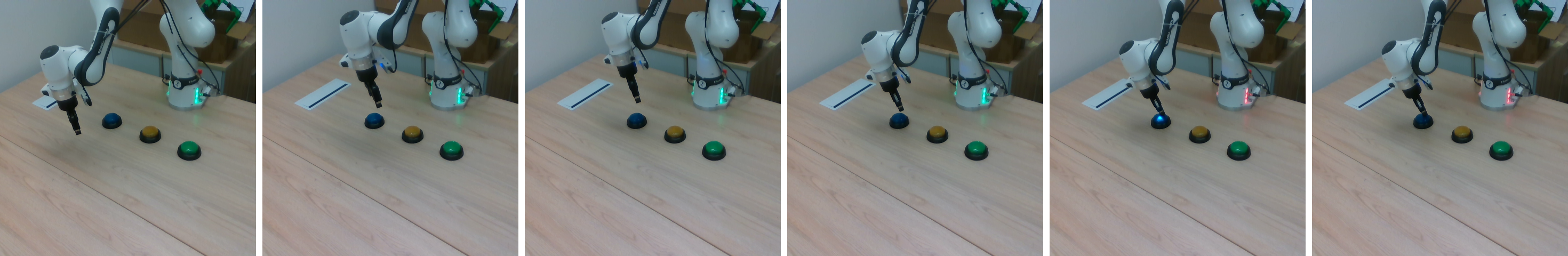}\vspace{1.5mm}\\
\bf Press the blue button.
\vspace{2mm}\\

\includegraphics[width=1\linewidth]{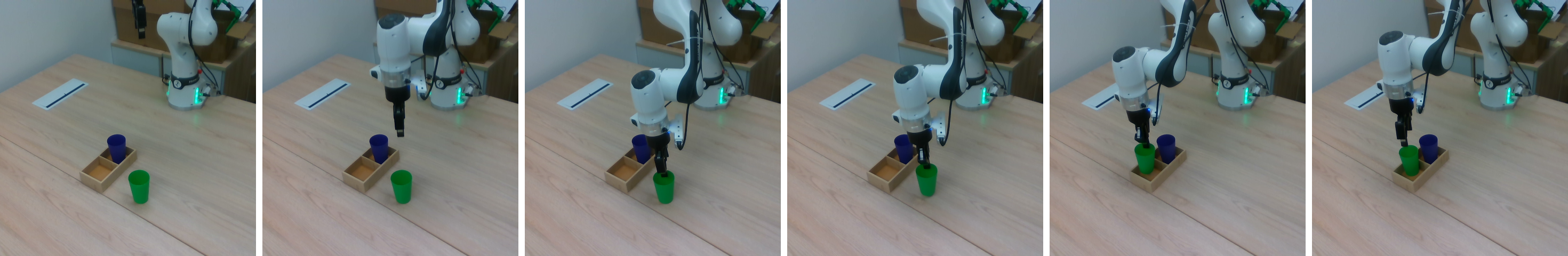}\vspace{1.5mm}\\
\bf Pick up the cup into the wood grid.
\vspace{2mm}

    \end{tabular}
    \caption{Data Samples of the Dataset on the Real Franka Emika Panda Robot Arm.}
    \label{fig:example1}
\end{figure*}

\begin{figure*}[h]
    \centering
    \begin{tabular}{c}
\includegraphics[width=1\linewidth]{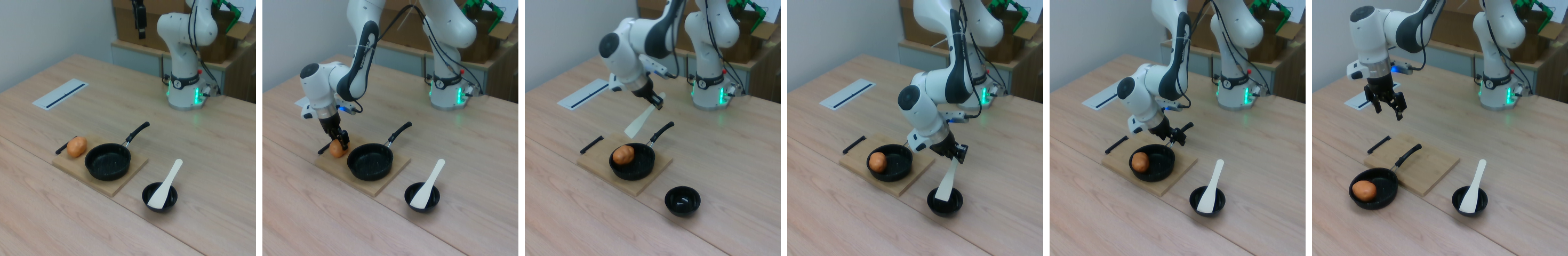}\vspace{1.5mm}\\
\bf Prepare the dishes.
\vspace{2mm}\\

\includegraphics[width=1\linewidth]{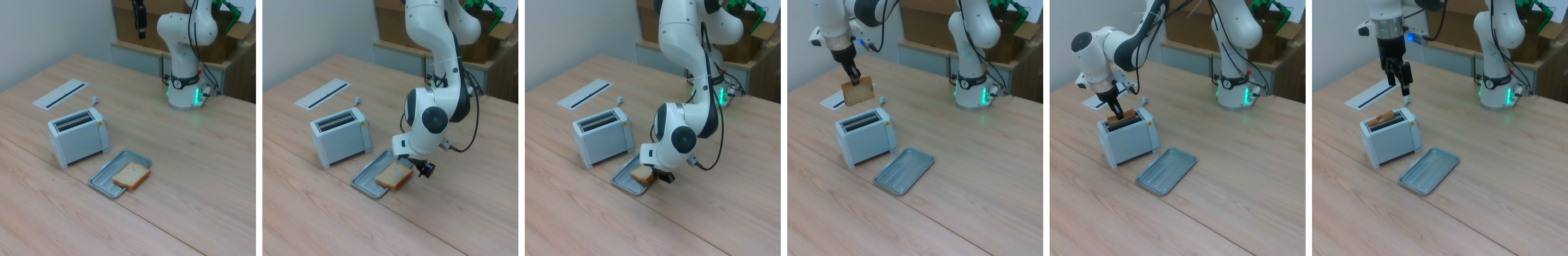}\vspace{1.5mm}\\
\bf Bake the bread.
\vspace{2mm}\\

\includegraphics[width=1\linewidth]{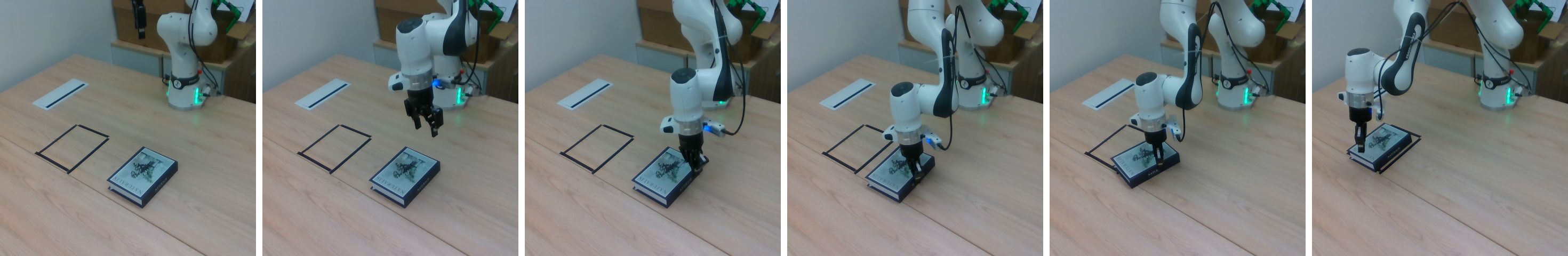}\vspace{1.5mm}\\
\bf Push the book into the black square.
\vspace{2mm}\\

\includegraphics[width=1\linewidth]{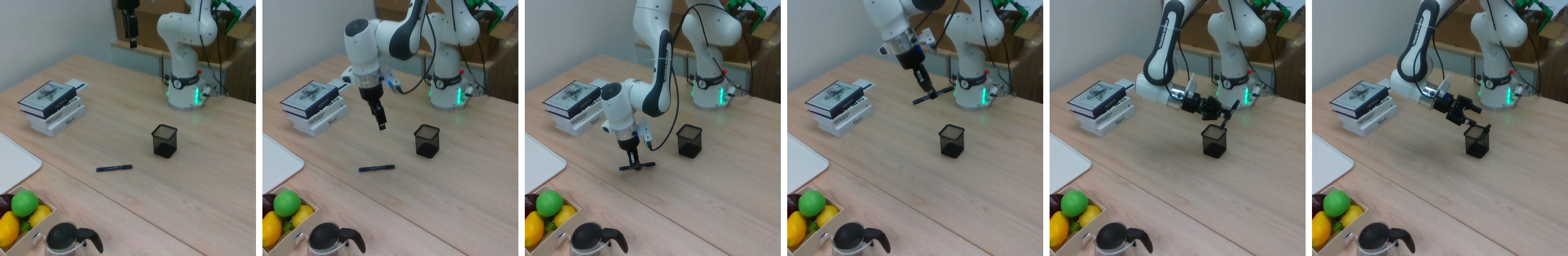}\vspace{1.5mm}\\
\bf Put the marker into the pen container.
\vspace{2mm}\\

\includegraphics[width=1\linewidth]{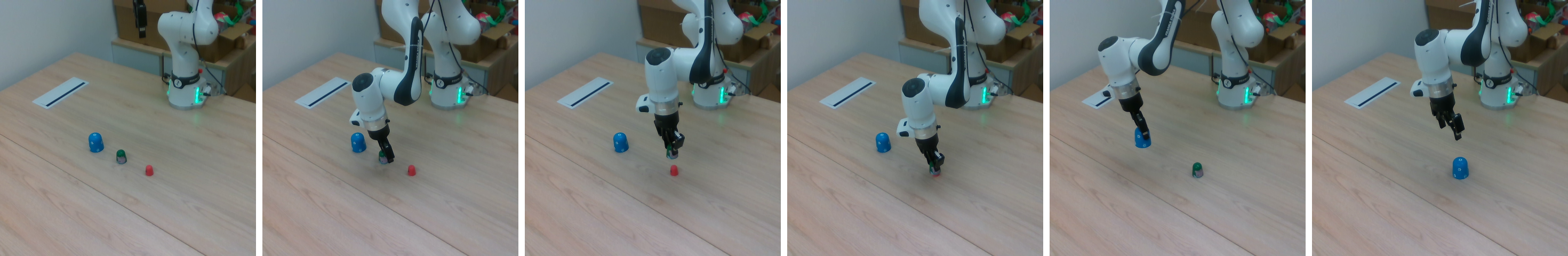}\vspace{1.5mm}\\
\bf Fold the nesting dolls together.
\vspace{2mm}
    \end{tabular}
    \caption{Data Samples of the Dataset on the Real Franka Emika Panda Robot Arm.}
    \label{fig:example2}
\end{figure*}

\begin{figure*}[h]
    \centering
    \begin{tabular}{c}
\includegraphics[width=1\linewidth]{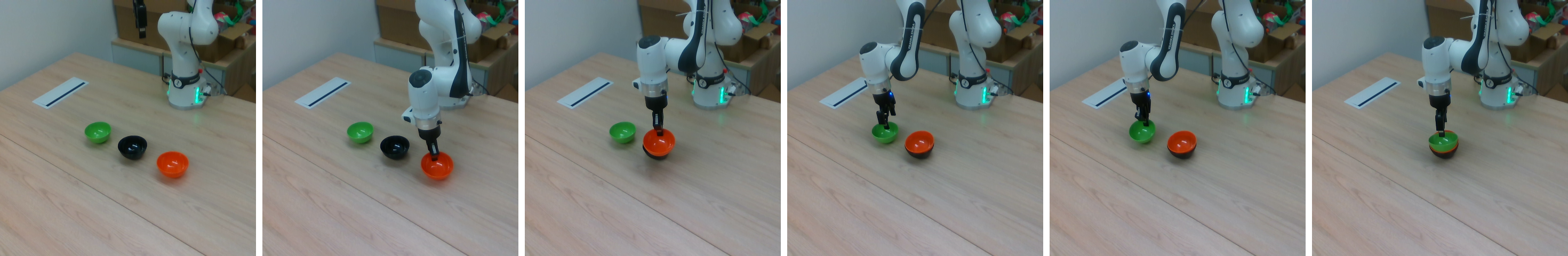}\vspace{1.5mm}\\
\bf Stack the bowls.
\vspace{2mm}\\

\includegraphics[width=1\linewidth]{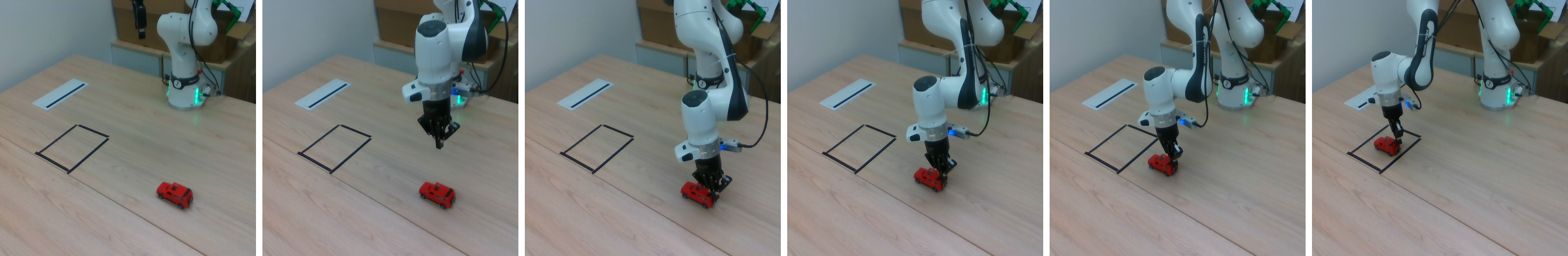}\vspace{1.5mm}\\
\bf Push the toy car into the black square.
\vspace{2mm}\\

\includegraphics[width=1\linewidth]{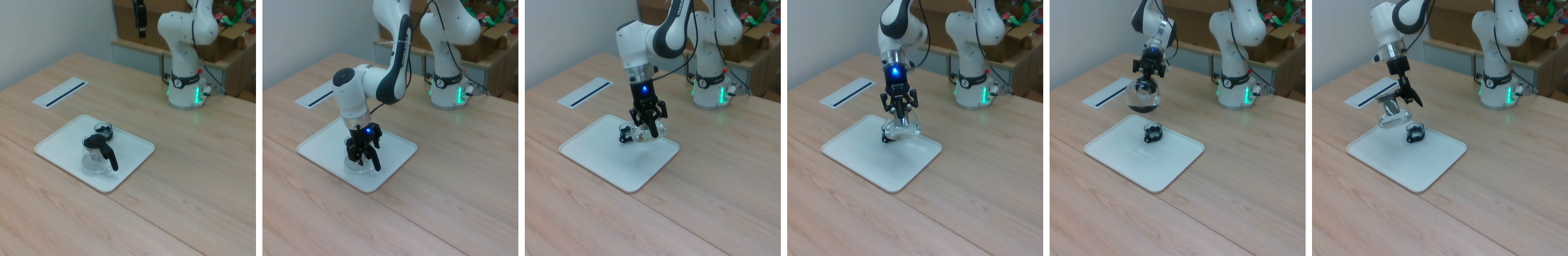}\vspace{1.5mm}\\
\bf Pour the water from the teapot to the cup.
\vspace{2mm}\\

\includegraphics[width=1\linewidth]{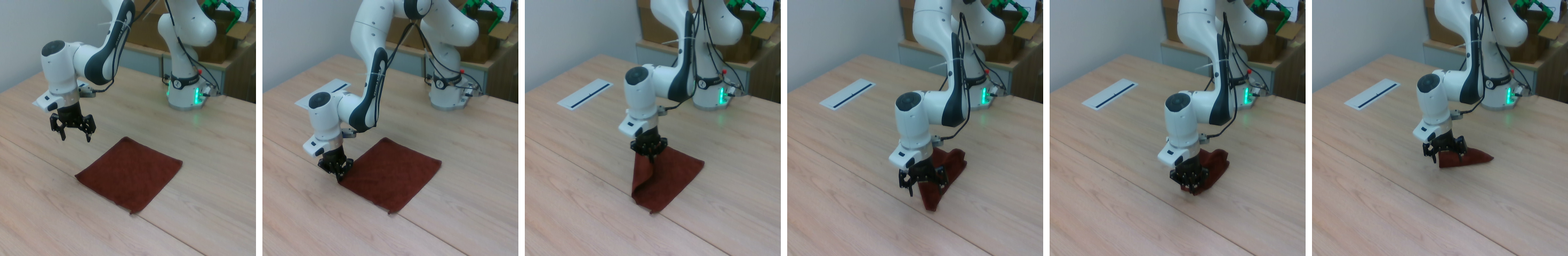}\vspace{1.5mm}\\
\bf Fold the towel.
\vspace{2mm}\\

\includegraphics[width=1\linewidth]{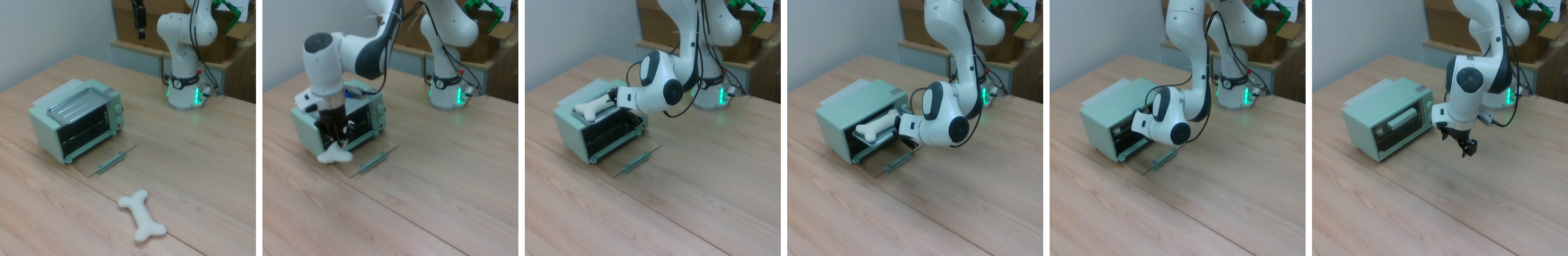}\vspace{1.5mm}\\
\bf Put the bone into the plate then put the plate intothe stove then close the stove.
\vspace{2mm}\\

    \end{tabular}
    \caption{Data Samples of the Dataset on the Real Franka Emika Panda Robot Arm.}
    \label{fig:example3}
\end{figure*}
To assess the real-world performance of our model, we conduct experiments on a Franka Emika Panda robotic arm using a privately collected dataset. We utilize two distinct camera setups for data collection: For Camera 1, we collect 15 tasks, labeled from Task 1 to Task 15. For Camera 2, we collect 8 tasks, labeled from Task 1 to Task 8. These task indices are consistently referenced throughout the Experiment Section in the paper. Representative examples from each task are shown in the Figure \ref{fig:example1}, Figure \ref{fig:example2} and Figure \ref{fig:example3}. 

In the following, we provide the full list of tasks along with their corresponding names and descriptions for reproducibility and clarity.
\begin{itemize}
    \item \textit{Task1: Pick up the carrot into the box.} This is a pick \& place task, pick up the carrot first then move it to the box successfully.
    \item \textit{Task2: Open the storage box.} This is a pick \& place task, the robot should precisely grasp the small handle of the box then open the box.
    \item \textit{Task3: Close the storage box.} This is a pick \& place task, grasp the small handle and move it to the top of the box.
    \item \textit{Task4: Press the (blue/red/yellow/green) button.} This is a press task. Press the center of the button and make the button flash.
    \item \textit{Task5: Pick up the cup into the wood grid.} This is a challenging pick \& place task due to the narrow rim of the cup and its high susceptibility to tipping over.
    \item \textit{Task6: Prepare the dishes.} This is a long horizon task. It can be divided in several parts: pick up the potatoes, move the potatoes into the pot, pick up the shovel and make a cook action above the pot, drop the shovel on the towel, push the pot off the chopping board.
    \item \textit{Task7: Bake the bread.} This is a challenging pick \& place task, pick up the bread and insert the bread into the toaster.
    \item \textit{Task8: Push the book into the black square.} This is a push task, push a large book into the selected area surrounding by the black tape.
    \item \textit{Task9: Put the marker into the pen container.} This is a pick \& place and large rotation task. Pick up the marker firstly then insert the marker into the pen container.
    \item \textit{Task10: Fold the nesting dolls together.} This is a long horizon task with hard grasping.
    \item \textit{Task11: Stack the bowls.} This is a long horizon task to stack three bowls together.
    \item \textit{Task12: Push the toy car into the black square.} This is a push task, pushing the small toy car precisely with the closed gripper.
    \item \textit{Task13: Pour the water from the teapot to the cup.} This is a pour task with hard grasping and large rotation.
    \item \textit{Task14: Fold the towel.} This is a soft-body manipulation task, fold the towel by grasp the corner of the towel and drag it to the suitable position.
    \item \textit{Task15: Put the bone into the plate then put the plate into the stove then close the stove.} This is a long horizon task. Pick up the bone, put it into the plate, grasp the plate and move it into the microwave,
    close the microwave.

\end{itemize}




\clearpage
\clearpage
\clearpage
\clearpage




\end{document}